\documentclass[11pt,a4paper]{article}

\usepackage[hyperref]{naaclhlt2019}

\usepackage{times}
\usepackage{latexsym}
\usepackage{url}
\usepackage{multirow}
\usepackage{graphicx}
\usepackage{subfig}
\usepackage{booktabs}

\aclfinalcopy 

\title{Bridging the Gap: Attending to Discontinuity in Identification of Multiword Expressions}

\newcommand*\samethanks[1][\value{footnote}]{\footnotemark[#1] }
\author{{\bf Omid Rohanian}\textsuperscript{$\dagger$}\thanks{*The first two authors contributed equally.} , {\bf Shiva Taslimipoor}\textsuperscript{$\dagger$}\samethanks, { \bf Samaneh Kouchaki}\textsuperscript{$\ddagger$}, {\bf Le An Ha}\textsuperscript{$\dagger$}, {\bf Ruslan Mitkov}\textsuperscript{$\dagger$}\\
\textsuperscript{$\dagger$}{Research Group in Computational Linguistics, University of Wolverhampton, UK}\\
\textsuperscript{$\ddagger$}{Institute of Biomedical Engineering, Department of Engineering Science, University of Oxford, UK}\\
{\tt  \{omid.rohanian, shiva.taslimi, l.a.ha, r.mitkov\}@wlv.ac.uk} \\
{\tt  samaneh.kouchaki@eng.ox.ac.uk} }

\date{}

\begin{document}

\maketitle
\begin{abstract}
We introduce a new method to tag Multiword Expressions (MWEs) 
using a linguistically interpretable language-independent deep learning architecture. 
We specifically target discontinuity, an under-explored aspect that poses a significant challenge to computational treatment of MWEs. Two neural architectures are explored: Graph Convolutional Network (GCN) and multi-head self-attention. GCN leverages dependency parse information, and self-attention attends to long-range relations. We finally propose a combined model that integrates complementary information from both, through a gating mechanism.
The experiments on a standard multilingual dataset for verbal MWEs show that our model outperforms the baselines not only in the case of discontinuous MWEs but also in overall F-score.\footnote{The code is available on \url{https://github.com/omidrohanian/gappy-mwes}.}     
\end{abstract}

\section{Introduction}
Multiword expressions (MWEs) are linguistic units composed of more than one word whose meanings cannot be fully determined by the semantics of their components \cite{Sag2002,baldwin2010multiword}. As they are fraught with syntactic and semantic idiosyncrasies, their automatic identification remains a major challenge \cite{constant2017multiword}. Occurrences of discontinuous MWEs are particularly elusive as they involve relationships between non-adjacent tokens (e.g. \textit{\textbf{put} one of the blue masks \textbf{on}}). 

While some previous 
studies disregard discontinuous MWEs \cite{legrand2016phrase}, others stress the importance of factoring them in 
\cite{Schneider14}. Using a CRF-based and a transition-based approach respectively, \citet{Moreau2018} and \citet{alsaied2017} try to capture discontinuous occurrences with help from dependency parse information. 
Previously explored neural MWE identification models \cite{Gharbieh2017} suffer from limitations in dealing with discontinuity, which can be attributed to their inherently sequential nature. More sophisticated architectures are yet to be investigated \cite{constant2017multiword}. 

Graph convolutional neural networks (GCNs) \cite{kipf2017semi} and attention-based neural sequence labeling \cite{tan2018deep} are methodologies suited for modeling non-adjacent relations and are hence adapted to MWE identification in this study. Conventional GCN \cite{kipf2017semi} uses a global graph structure 
for the entire input. We modify it such that
GCN filters convolve nodes of dependency parse tree on a per-sentence basis. Self-attention, on the other hand, learns representations by relating different parts of the same sequence. Each position in a sequence is linked to any other position with $O(1)$ operations, minimising maximum path (compared to RNN's $O(n)$) which facilitates gradient flow and makes it theoretically well-suited for learning long-range dependencies \cite{vaswani2017attention}. 

The difference in the two approaches motivates our attempt to incorporate them into a hybrid model with an eye to exploiting their individual strengths. 
Other studies that used related syntax-aware methods in sequence labeling include \citet{marcheggiani2017encoding} and \citet{strubell2018} where GCN and self-attention were separately applied to semantic role labelling. 

Our contribution in this study, is to show for the first time, how GCNs can be successfully applied to MWE identification, especially to tackle discontinuous ones. Furthermore, we propose a novel architecture that integrates GCN with self-attention outperforming state-of-the-art. The resulting models not only prove superior to existing methods in terms of overall performance but also are more robust in handling cases with gaps.


\section{Methodology}
To specifically target discontinuity, we explore two mechanisms both preceding a Bi-LSTM: 1) a GCN layer to act as a syntactic ngram detector, 2) an attention mechanism to learn long-range dependencies.

\subsection{Graph Convolution as Feature Extraction}
\label{sec:graphconv}
Standard convolutional filters act as sequential ngram detectors \cite{kim2014convolutional}. Such filters might prove inadequate in modeling complex language units like discontinuous MWEs. One way to overcome this problem is to consider non-sequential relations by attending to syntactic information in parse trees through the application of GCNs.

GCN is defined as a directed multi-node graph $G({V},{E})$ where ${v}_i \in {V}$ and $({v}_i, {r}, {v}_j) \in {E}$ are entities (words) and edges (relations) respectively. By defining a vector ${x}_{v}$ as the feature representation for the word ${v}$, the convolution equation in GCN can be defined as a non-linear activation function $f$ and a filter ${W}$ with a bias term $b$ as:
\begin{equation}
{c} = {f}(\sum_{i \in {r}({v})}{W} {x}_i + b) 
\label{eq:conv}
\end{equation}
where ${r}({v})$ shows all words in relation with the given word ${v}$ in a sentence, and ${c}$ represents the output of the convolution. 

Following \newcite{kipf2017semi} and \newcite{schlichtkrull2017}, we represent graph relations using adjacency matrices as mask filters for inputs. We derive associated words from the dependency parse tree of the target sentence. Since we are dealing with a sequence labelling task, there is an adjacency matrix representing relations among words (as nodes of the dependency graph) for each sentence. We define the sentence-level convolution operation with filter ${W_s}$ and bias $b_s$  as follows:
\begin{equation}
{C}_s = {f}({W}_s {X}^T {A} + b_s) 
\label{eq:gconv}
\end{equation}  
where ${X}$, ${A}$, and ${C}$ are representation of words, adjacency matrix, and the convolution output, all at the level of sentence. The above formalism considers only one relation type, while depending on the application, multiple relations can be defined. 

\citet{kipf2017semi} construct separate adjacency matrices corresponding to each relation type and direction. Given the variety of dependency relations in a parse tree (e.g. obj, nsubj, advcl, conj, etc), and per-sentence adjacency matrices, we would end up with an over-parametrised model in a sequence labeling task. In this work, we simply treat all relations equally, but consider only three types of relations: 1) the head to the dependents, 2) the dependents to the head, and 3) each word to itself (self-loops). The final output is obtained by aggregating the outputs from the three relations. 
 
\subsection{Self-Attention}
\label{sec:att}
Attention \cite{bahdanau2014neural} helps a model address the most relevant parts of a sequence through weighting. As attention is designed to capture dependencies in a sequence regardless of distance, it is complementary to RNN or CNN models 
where longer distances pose a challenge. 
In this work we employ multi-head self-attention with a weighting function based on scaled dot product which makes it fast and computationally efficient. 

Based on the formulation of Transformer by \citet{vaswani2017attention}, in the encoding module an input vector ${x}$ is mapped to three equally sized matrices ${K}$, ${Q}$, and ${V}$ (representing key, query and value) and the output weight matrix is then computed as follows: 
\begin{equation} \label{eq:scaledDotProduct}
\mbox{Att}({Q},{K},{V}) = \mbox{softmax}(\frac{{Q{K}}^{T}}{\sqrt{d}}){V}
\end{equation}
The timing signal required for the self-attention to work is already contained in the preceding CNN layers alleviating the need for position encoding.

\subsection{Model Architecture}
\begin{figure}[ht]
\centering
\includegraphics[width=0.50\textwidth]{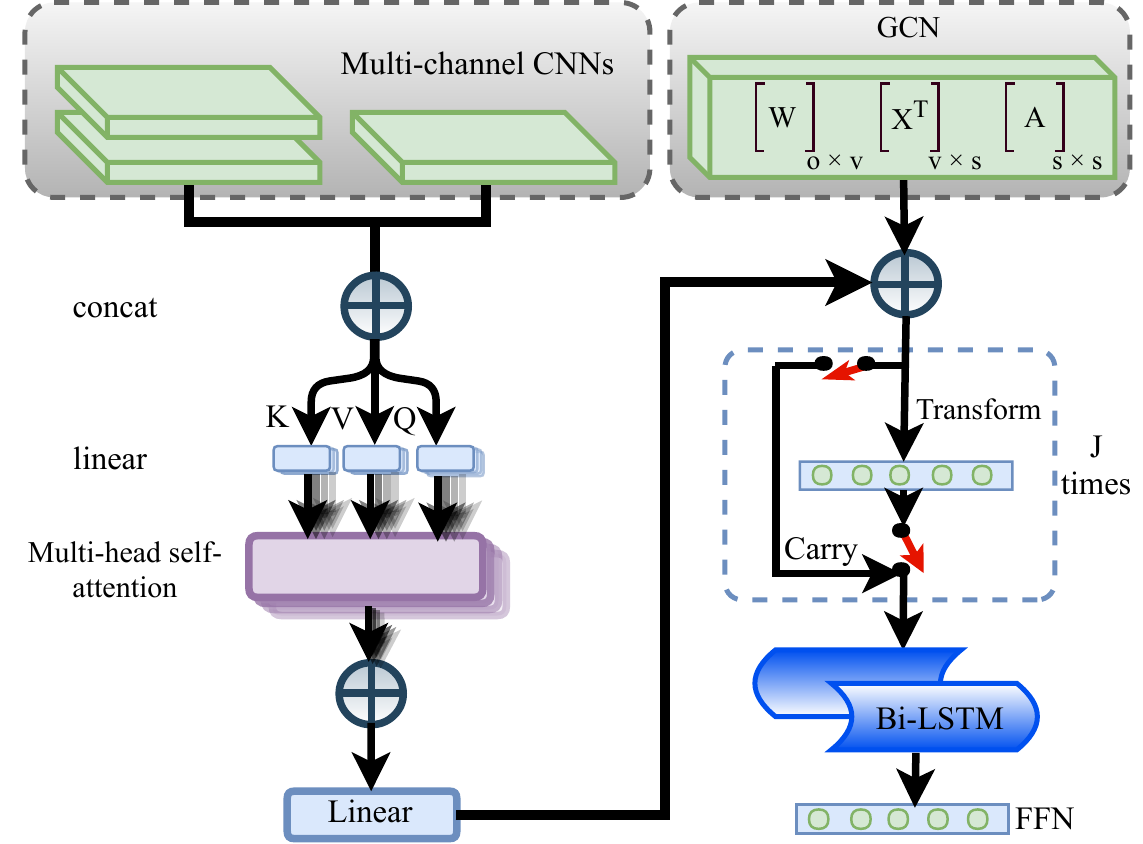}
\caption{A hybrid sequence labeling approach integrating GCN (o: output dimension; v: word vectors dimension; s: sentence length) and Self-Attention.}
\label{fig:diag}
\end{figure}

The overall scheme of the proposed model, composed of two parallel branches, is depicted in Figure \ref{fig:diag}. We employ multi-channel CNNs as the step preceding self-attention. One channel is comprised of two stacked 1D CNNs and the other is a single 1D CNN. 
After concatenation and batch normalisation, a multi-head self attention mechanism is applied (Section \ref{sec:att}).

Parallel to the self-attention branch, GCN learns a separate representation (Section \ref{sec:graphconv}). 
Since the GCN layer retains important structural information and is sensitive to positional data from the syntax tree, we consider it as a position-based approach.
On the other hand, the self-attention layer is intended to capture long-range dependencies in a sentence. It relates elements of the same input through a similarity measure irrespective of their distance. We therefore regard it as a content-based approach. As these layers represent different methodologies, 
we seek to introduce a model that combines their complementary traits in our particular task.
\vspace*{2mm}

\noindent
\textbf{Gating Mechanism}. Due to the considerable overlap between the GCN and self-attention layers, a naive concatenation introduces redundancy which significantly lowers the learning power of the model. To effectively integrate the information, 
we design a simple gating mechanism using feed-forward highway layers \citep{srivastava2015highway} which learn to regulate information flow in consecutive training epochs. Each highway layer consists of a Carry ($Cr$) and a Transform ($Tr$) gate which decide how much information should pass or be modified. 
For simplicity $Cr$ is defined as $1 - Tr$. We apply a block of $J$ stacked highway layers (the section inside the blue dotted square in Figure \ref{fig:diag}). 
Each layer regulates its input ${x}$ using the two gates and a feedforward layer $H$ as follows:
\begin{equation}
y = Tr\odot H+(1-Tr)\odot {x}
\end{equation}
where $\odot$ denotes the Hadamard product and $Tr$ is defined as $\sigma ({W}_{Tr}{x} + b_{Tr})$. We set $b_{Tr}$ to a negative number to reinforce carry behavior which helps the model learn temporal dependencies early in the training.

Our architecture bears some resemblance to \citet{marcheggiani2017encoding} and \citet{zhang2018graph} in its complementary view of GCN and BiLSTM. However there are some important differences. In these works, BiLSTM is applied prior to GCN in order to encode contextualised information and to enhance the teleportation capability of GCN. \citet{marcheggiani2017encoding} stack a few BiLSTM layers with the idea that the resulting representation would enable GCN to consider nodes that are multiple hops away in the input graph. \citet{zhang2018graph} use a similar encoder, however the model employs single BiLSTM and GCN layers, and the graph of relations is undirected.

In our work, we use pre-trained contextualised embeddings that already contain all the informative content about word order and disambiguation. We put BiLSTM on top of GCN, in line with how CNNs are traditionally applied as feature generating front-ends to RNNs. Furthermore, \citet{marcheggiani2017encoding} use an edge-wise gating mechanism in order to down-weight uninformative syntactic dependencies. This method can mitigate noise when parsing information is deemed noisy, however in \citet{zhang2018graph} it caused performance to drop. Given our low-resource setting, in this work we preferred not to potentially down-weight contribution of individual edges, therefore treating them equally. We rely on gating as the last step when we combine GCN and self-attention.         

\section{Experiments}

\noindent
\textbf{Data}. We experiment with datasets from the shared task on automatic identification of verbal Multiword Expressions \cite{ramisch2018edition}. The datasets are tagged for different kinds of verbal MWEs including idioms, verb particle constructions, and light verb constructions among others. We focus on annotated corpora of four languages: French (FR), German (DE), English (EN), and Persian (FA) due to their variety in size and proportion of discontinuous MWEs. Tags in the datasets are converted to a variation of IOB  which includes the tags \texttt{B} (beginning of MWEs), \texttt{I} (other components of MWEs), and \texttt{O} (tokens outside 
MWEs), with the addition of \texttt{G} for arbitrary tokens in between the MWE components e.g. $make_{\texttt{[B]}}$ $important_{[\texttt{G}]}$ $decisions_{[\texttt{I}]}$.

\vspace*{2mm}
\noindent
\textbf{ELMo}. In our experiments, we make use of ELMo embeddings \cite{Peters2018} which are 
contextualised and token-based as opposed to type-based word representations like \texttt{word2vec} or \texttt{GLoVe} where each word type is assigned a single vector. 
Token-based embeddings better reflect the syntax and semantics of each word in its context compared to traditional type-based ones. We use the implementation by \citet{che-EtAl:2018:K18-2} to train ELMo embeddings on our data. 

\vspace*{2mm}
\noindent
\textbf{Validation}. In the validation phase, we start with a strong baseline which is a CNN + Bi-LSTM model based on the top performing system in the VMWE shared task \cite{taslimipoor2018shoma}. Our implemented baseline differs in that we employ ELMo rather than \texttt{word2vec} resulting in a significant improvement.
We perform hyper-parameter optimisation and make comparisons among our systems, including  GCN + Bi-LSTM (GCN-based), CNN + attention + Bi-LSTM (Att-based), and their combination using a highway layer (H-combined) in Table \ref{results_dev}.

\begin{table}[t!]
\small
\begin{center}
\setlength{\tabcolsep}{0.3em}
\begin{tabular}{|ll|c|c|c c c c |}
\hline
& & \multicolumn{2}{c|}{All} &  \multicolumn{4}{c|}{Discontinuous} \\
& & Token- & MWE- &  \multicolumn{4}{c|}{} \\
& & \multicolumn{1}{c|}{based} & \multicolumn{1}{c|}{based} &   & \multicolumn{3}{c|}{MWE-based}\\
L & model  & F  & F & \% & P & R & F\\ 
\hline
\multirow{4}{*}{EN} & baseline  &  41.37 &  {35.38} & \multirow{4}{*}{32} & 24.44 & 10.48 & 14.67 \\
\multirow{2}{*}{} & GCN-based  & 39.78 &  {39.11} & &   39.53 & 16.19 & 22.97\\
& Att-based &  33.33 & 31.79 & & 46.88 & 14.29 & 21.90     \\
& H-combined &  {41.63} &  \textbf{40.76} & & 63.33 & 18.10 & \textbf{28.15} \\
\hline
\multirow{4}{*}{DE} & baseline   & 62.27 &  {57.17} & \multirow{4}{*}{43} & 69.50 & 45.37 & 54.90 \\
\multirow{2}{*}{} & GCN-based  &  {65.48} &  \textbf{61.17} & & 65.19 & 47.69 & 55.08 \\
 & Att-based & 61.20 & 58.19 & & 67.86 & 43.98 & 53.37  \\
& H-combined &  {63.80} &  {60.71} & & 68.59 & 49.54 & \textbf{57.53} \\ 
\hline
\multirow{4}{*}{FR} & baseline & {76.62} & {72.16} & \multirow{4}{*}{43} & 75.27 & 52.04 & 61.54 \\
\multirow{2}{*}{} & GCN-based  &  {79.59} & {75.15} & & 79.58 & 56.51 &  66.09 \\
& Att-based & 78.21 & 74.23 & & 71.49 & 60.59 & 65.59      \\
& H-combined & {80.25} & \textbf{76.56} & & 77.94 & 59.11 & \textbf{67.23} \\
\hline
\multirow{4}{*}{FA} & baseline  & {88.45} & {86.50} & \multirow{4}{*}{14} & 67.76 & 55.88 & 61.29 \\
\multirow{2}{*}{} & GCN-based  &  {87.78} &  {86.42} & & 78.72 & 54.41 &  64.35 \\
& Att-based & 87.55 & 84.20 & & 62.32 & 63.24 & 62.77     \\
& H-combined & {88.76} & \textbf{87.15} & & 75.44 & 63.24 & \textbf{68.80}  \\
\hline
\end{tabular}
\caption{\label{results_dev} Model performance (P, R and F) for development sets for all MWE and only discontinuous ones (\%: proportion of discontinuous MWES) }
\vspace*{-\baselineskip}
\end{center}
\end{table}

\section{Evaluation and Results}
Systems are evaluated using two types of precision, recall and F-score measures: strict MWE-based scores (every component of an MWE should be correctly tagged to be considered as true positive), and token-based scores (a partial match between a predicted and a gold MWE would be considered as true positive). We report results for all MWEs as well as discontinuous ones specifically.

\begin{table*}[t]
\begin{center}
\begin{tabular}{|l|c || c||c|| c|| c|c|c|c|}
\hline
 & \multicolumn{4}{c|}{All $|$ Discontinuous} \\
 & EN & DE & FR & FA 
 \\
\hline
baseline 
& 33.01 $|$ 16.53 & 54.12 $|$ 53.94 & 67.66 $|$ 58.70 & \textbf{81.62} $|$ 61.73 
\\
\hline
GCN-based 
& 36.27 $|$ \textbf{24.15} & 56.96 $|$ 54.87 & 70.79 $|$ 59.95 & 81.00 $|$ \textbf{62.35}  
\\
H-combined 
& \textbf{41.91} $|$ 22.73 & \textbf{59.29} $|$ \textbf{55.00} 
& \textbf{70.97} $|$ \textbf{63.90} & 80.04 $|$ 61.90 
\\
\hline
ATILF-LLF 
& 31.58 $|$ 09.91 & 54.43 $|$ 40.34 & 58.60 $|$ 51.96 & 77.48 $|$ 53.85 
\\
SHOMA 
& 26.42 $|$ 01.90  & 48.71 $|$ 40.12 & 62.00 $|$ 51.43 & 78.35 $|$ 56.10 
\\
\hline
\end{tabular}
\end{center}
\caption{\label{testResults} Comparing the performance of the systems on test data in terms of MWE-based F-score }
\end{table*}

According to Table \ref{results_dev}, GCN-based outperforms Att-based and they both outperform the strong baseline in terms of MWE-based F-score in three out of four languages. Combining GCN with attention using highway networks results in further improvements for EN, FR and FA. The H-combined model consistently exceeds the baseline for all languages. As can be seen in Table \ref{results_dev}, GCN and H-combined models each show significant improvement with regard to discontinuous MWEs, regardless of the proportion of such expressions.

In Table \ref{testResults} we show the superior performance (in terms of MWE-based F-score) of our top systems on the test data compared to the baseline and state-of-the-art systems, namely, ATILF-LLF \cite{alsaied2017} and SHOMA \cite{taslimipoor2018shoma}. GCN works the best for discontinuous MWEs in EN and FA, while H-combined outperforms based on results for all MWEs except for FA. The findings are further discussed in Section \ref{sec:discussion}.

\section{Discussion and Analysis}
\label{sec:discussion}
\begin{figure}[ht]
\begin{minipage}{.49\linewidth}
\centering
\includegraphics[scale=0.29]{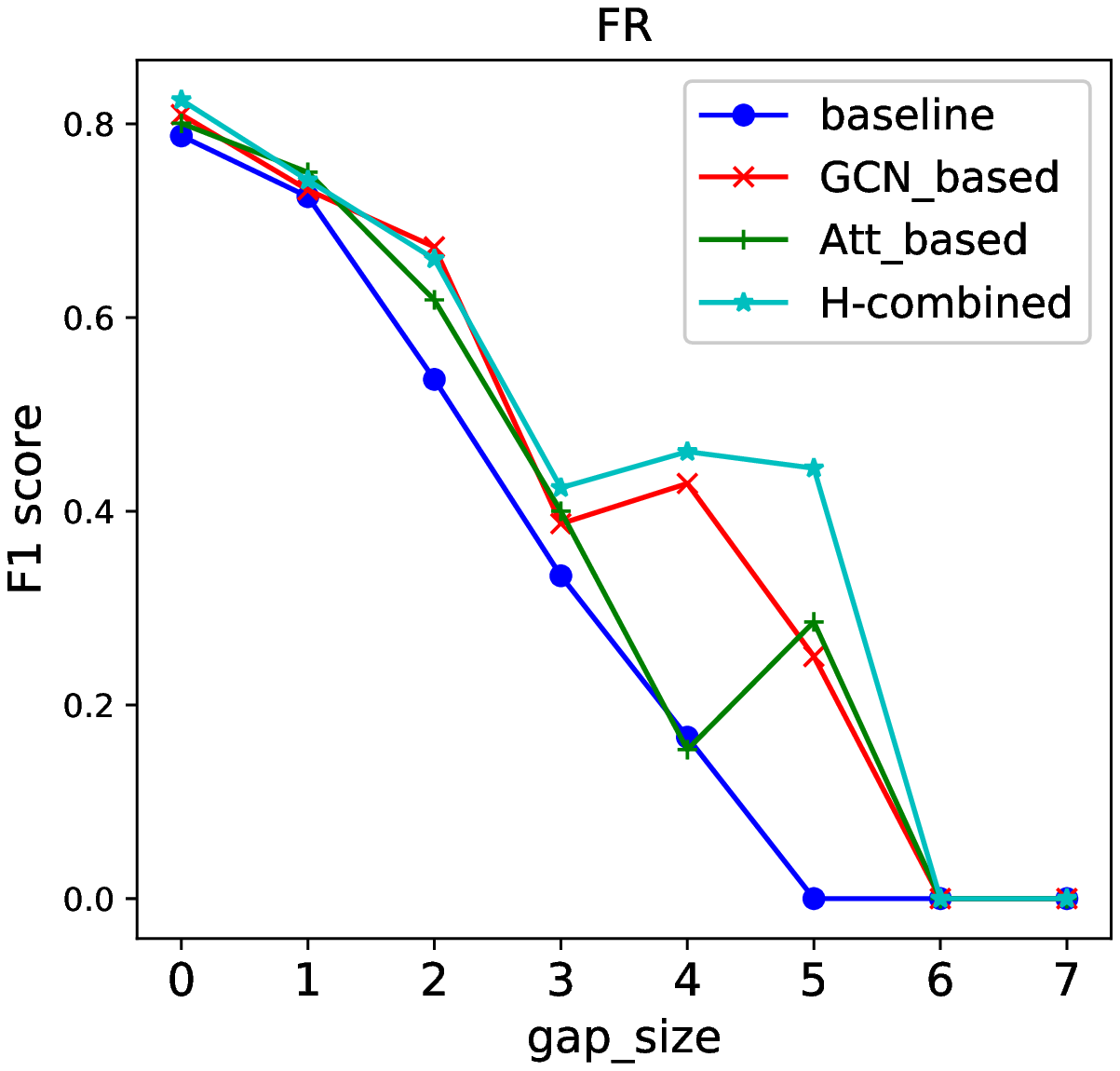} 
\end{minipage}
\begin{minipage}{.49\linewidth}
\centering
\includegraphics[scale=0.29]{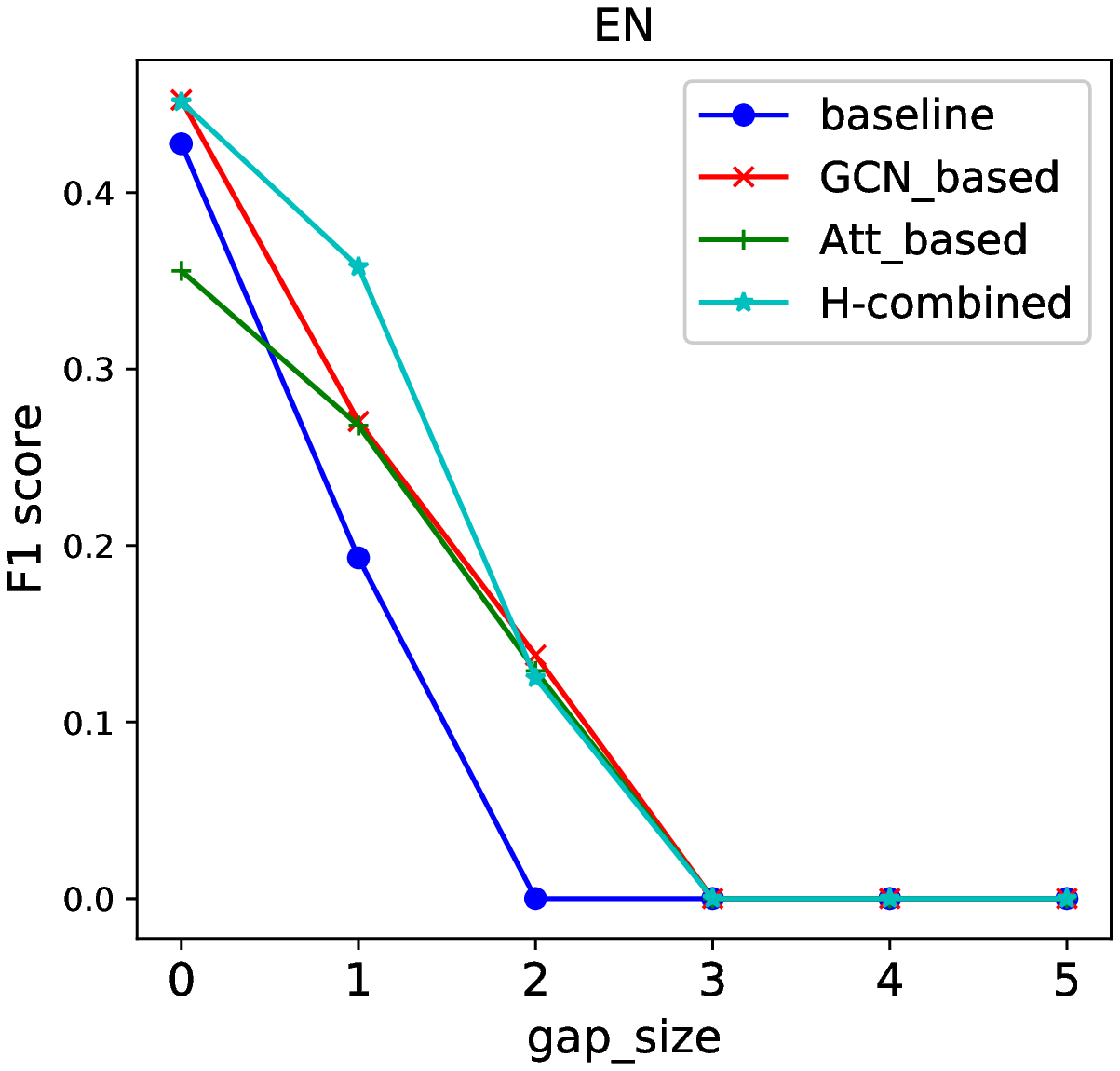} 
\end{minipage}

\caption{Model performance given different gap sizes}
\label{fig:discountinuity}
\end{figure}

The overall results confirm our assumption that a hybrid architecture can mitigate errors of individual models and bolster their strengths. 
To demonstrate the effectiveness of the models in detecting discontinuous MWEs, in Figure \ref{fig:discountinuity} we plot their performance for FR and EN given a range of different gap sizes. As an ablation study, we show the results for the baseline, GCN-based, Att-based only, as well as H-combined models. 
GCN and Att-based models each individually outperform the baseline, and the combined model clearly improves the results further. 

\begin{figure}[h!]
\centering
\includegraphics[width=0.48\textwidth]{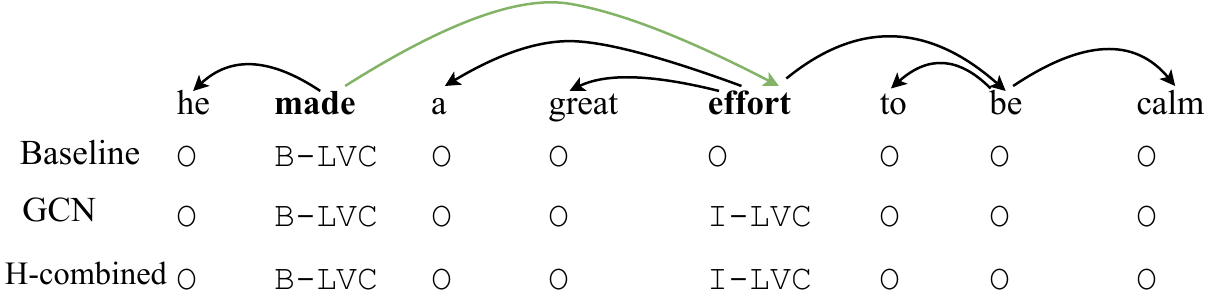}
\noindent
\caption{Sample sentence with a discontinuous occurrence of an English MWE, \textit{make an effort}.}
\label{fig:gappyen}
\end{figure}

The example in Figure \ref{fig:gappyen} taken from the English dataset demonstrates the way GCN considers relations between non-adjacent tokens in the sentence. Our baseline is prone to disregarding these links. Similar cases captured by both GCN and H-combined (but not the baseline) are \textit{\textbf{take} a final \textbf{look}}, \textit{\textbf{picked} one \textbf{up}}, and \textit{\textbf{cut} yourself \textbf{off}}.

\begin{figure}[ht]
\centering
\includegraphics[width=0.48\textwidth]{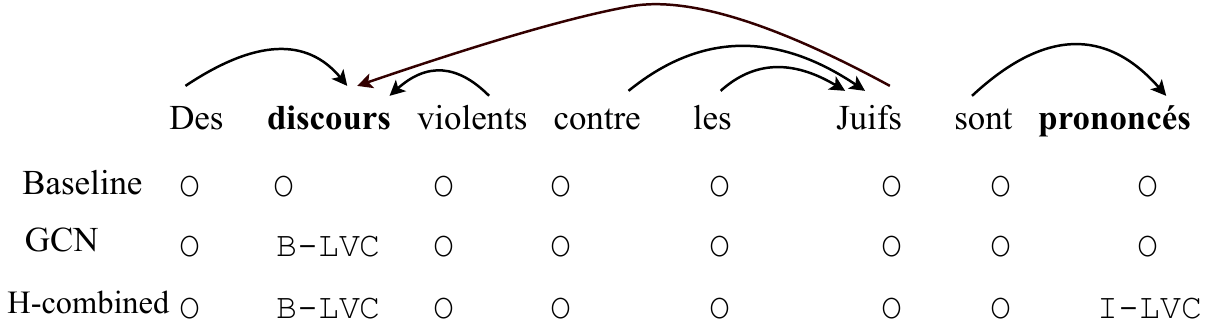}
\noindent
\caption{Example sentence with a discontinuous occurrence of a French MWE, \textit{prononcer un discours} `to make a speech'.}
\label{fig:gappyfr}
\vspace*{-\baselineskip}
\end{figure}

In more complicated constructs where syntactic dependencies might not directly link all constituents, GCN alone is not always conducive to optimal performance. In Figure \ref{fig:gappyfr}, the French sentence is in the passive form and MWE parts are separated by 5 tokens. This is an MWE skipped by GCN but entirely identified by the H-combined model. 

It is important to note that model performance is sensitive to factors such as percentage of seen expressions and variability of MWEs \citep{pasquer2018towards}. In FA for instance, 67\% of the MWEs in the test set are seen at training time, making them easy to be captured by the baseline \citep{taslimipoorPMWE}. Furthermore, only 21\% of MWEs in FA and 15\% in EN are discontinuous as opposed to 44\% in FR and 38\% in DE. In this case, a sequential model can already learn the patterns with high accuracy and the potential of a GCN and self-attention is not fully exploited. 

Also in DE, a sizable portion of MWEs are verbal idioms (VIDs) which are known for their lexico-syntactic fixedness and prevalence of tokens that lack a standalone meaning and occur only in a limited number of contexts (also known as cranberry words). Furthermore, MWEs in the Persian dataset are all Light Verb Constructions (LVCs), which can be modelled using lexical semantic templates \citep{megerdoomian2004semantic}. For such MWEs, our models compete with strong sequential baselines. 







\section{Conclusion and Future Work}
In this paper, we introduced the application of GCN and attention mechanism to identification of verbal MWEs and finally proposed and tested a hybrid approach integrating both models. Our particular point of interest is discontinuity in MWEs which is an under-explored area. All the individual and combined models outperform state-of-the-art in all considered criteria. 
In future, we will further develop our system using structured attention \citep{kim2017structured} and try to improve the accuracy of parsers in multi-tasking scenarios.


\bibliography{naaclhlt2019}
\bibliographystyle{acl_natbib}

\end{document}